\documentclass[letterpaper, 10 pt, conference]{ieeeconf} 
\IEEEoverridecommandlockouts 
\usepackage{microtype}
\usepackage{graphicx}
\usepackage{caption}
\usepackage{subcaption}
\usepackage{booktabs} % for professional tables
\usepackage{hyperref}
\usepackage[dvipsnames]{xcolor}
\usepackage{amsmath}
\usepackage{amsfonts}
\usepackage{amssymb}
\usepackage{cleveref}
\usepackage{multirow}
\usepackage{bm}
\usepackage{cite}
\usepackage[ruled, vlined]{algorithm2e}
\usepackage[nodisplayskipstretch]{setspace}
\usepackage{etoolbox}

\DeclareMathOperator*{\argmin}{arg\,min}

\newcommand{\inner}[2]{\langle#1, #2\rangle}

\newcommand{\Expect}[2]{\mathbb{E}_{#1}\Big[#2\Big]}

\newcommand{\real}{\mathbb{R}}

\newcommand{\relu}{\text{ReLU}}

\newcommand{\hrulethick}{\specialrule{0.1em}{0em}{0em}}

\SetKwInput{KwParam}{Parameters}
\SetKwFor{ParallelFor}{for}{do in parallel}{end}

\providetoggle{beginFlag}
\settoggle{beginFlag}{true}

\SetAlgoSkip{customskip}

\title{\LARGE \bf
Learning to Optimize in Model Predictive Control
}

\author{Jacob Sacks$^{*}$ and Byron Boots$^{*}$% <-this % stops a space
\thanks{$^{*}$University of Washington, Seattle WA 98105, USA. 
        {\tt\small \{jsacks6, bboots\}@cs.washington.edu}}%
}

\begin{document}

\maketitle

\begin{abstract}
Sampling-based Model Predictive Control (MPC) is a flexible control framework that can reason about non-smooth dynamics and cost functions.
Recently, significant work has focused on the use of machine learning to improve the performance of MPC, often through learning or fine-tuning the dynamics or cost function.
In contrast, we focus on learning to \emph{optimize} more effectively. In other words, to improve the update rule within MPC. 
We show that this can be particularly useful in sampling-based MPC, where we often wish to minimize the number of samples for computational reasons. 
Unfortunately, the cost of computational efficiency is a reduction in performance; fewer samples results in noisier updates.
We show that we can contend with this noise by learning how to update the control distribution more effectively and make better use of the few samples that we have.
Our learned controllers are trained via imitation learning to mimic an expert which has access to substantially more samples.
We test the efficacy of our approach on multiple simulated robotics tasks in sample-constrained regimes and demonstrate that our approach can outperform a MPC controller with the same number of samples.
\vspace{-3ex}
\end{abstract}

\section{Introduction}
Model Predictive Control (MPC) is a powerful, practical tool for solving sequential decision problems on real-world systems. % in highly stochastic and non-stationary environments. 
%The approach can be viewed as a receding horizon algorithm that achieves robustness through continual optimization. The general idea is to use an approximate model to plan for a finite horizon, execute the first step, and then re-optimize given state feedback.
MPC has been successfully used in a variety of tasks including autonomous helicopter aerobatics \cite{abbeel2010autonomous}, aggressive off-road driving \cite{williams2016aggressive, williams2017information, wagener2019online}, manipulation \cite{kumar2016optimal,bhardwaj2021fast}, and humanoid robot locomotion \cite{erez2013integrated}. 
%
%
%Gradient-based approaches to MPC require that the cost and dynamics functions are differentiable, which can be restrictive.
%In contrast, sampling-based methods are more flexible, as they allow for generic cost and dynamics functions.
Recent work~\cite{wagener2019online} has shown that many popular MPC algorithms can be unified through the generic framework of dynamic mirror descent (DMD) \cite{hall2013dynamical}, a first-order online learning algorithm.
%From this perspective, many sampling-based MPC algorithms are effectively performing a gradient-based update with a Monte Carlo approximation of the gradient.
This perspective provides an opportunity to improve performance of existing algorithms by drawing on powerful optimization techniques. 
Most modern approaches to optimization use fixed update rules tailored to specific classes of problems.
%For first-order methods, this includes techniques such as momentum \cite{nesterov27method, qian1999momentum} or scaling updates based on the gradient history \cite{duchi2011adaptive, zeiler2012adadelta, kingma2014adam, reddi2019convergence}.
Recently, research has explored \emph{learning} to optimize \cite{chen2021learning}, where the update rule is specified by a function approximator, such as a neural network, that can improve optimization performance with experience.
%These methods learn the optimizer either by differentiating through the optimization process using gradient descent \cite{andrychowicz2016learning, ravi2016optimization, lv2017learning, chen2017learning, chen2020training} or using reinforcement learning, where the reward is optimizer performance \cite{li2016learning, wang2016learning, li2017learning, bello2017neural}.

In this work, we leverage the optimization perspective of sampling-based MPC and adopt the learning-to-optimize framework in order to improve the update rule.
This is in contrast to most existing learning-based approaches to MPC, which either focus on learning a good dynamics model \cite{kocijan2004gaussian, lenz2015deepmpc, fu2016one, williams2017information, nagabandi2018neural, chua2018deep, hafner2019learning}, introducing a learned cost-shaping term into the objective \cite{tamar2017learning}, coupling MPC with a learned value function \cite{zhong2013value,rosolia2017learning,lowrey2018plan,bhardwaj2020information}, or learning a good warm-start for MPC to refine \cite{wang2019exploring}.
Other approaches have focused on performing MPC with a learned latent space of high-dimensional observations \cite{wahlstrom2015pixels, watter2015embed, finn2017deep, banijamali2018robust, ebert2018visual, ha2019adaptive, hafner2019learning}, low-level skills \cite{sharma2019dynamics}, or controls such that sampling because more efficient \cite{amos2020differentiable}.
Another promising avenue has explored differentiating through optimal controllers or planners to learn components of the optimization pipeline.
These methods propose to learn or fine-tune the dynamics and cost functions of the controllers \cite{tamar2017learning, amos2018differentiable, karkus2017qmdp, okada2017path, okada2018acceleration, pereira2018mpc, amos2020differentiable} or parameters of differentiable planners \cite{tamar2016value, srinivas2018universal, yu2019unsupervised, bhardwaj2020differentiable} end-to-end.
%Another promising avenue has explored differentiating through optimal controllers \cite{tamar2017learning, amos2018differentiable, karkus2017qmdp, okada2017path, okada2018acceleration, pereira2018mpc, amos2020differentiable} or planners \cite{tamar2016value, srinivas2018universal, yu2019unsupervised, bhardwaj2020differentiable} to learn components of the optimization pipeline end-to-end.
%By jointly learning the dynamics model along with the cost function, these approaches can optimize for our actual performance, rather than surrogate losses, such as data likelihood.
%Unlike our work, all of these approaches leave the hand-designed update rule fixed.
%These are orthogonal objectives and potentially could be combined to achieve even better performance.
Compared with these approaches, we fix the dynamics and cost function and instead focus on improving the optimization process.

%One of the typical goals in learning-to-optimize is to speed up convergence to a local optima.
%However, converging rapidly in MPC is often, counter-intuitively, \emph{not} desirable.
%From the DMD perspective on MPC, the optimization objective is changing over time, and converging to a solution at one time step may lead to difficulties optimizing the objective at the next. 
For many practical sampling-based MPC algorithms, the primary challenge is finding a good trade-off between speed and accuracy. 
Sampling-based MPC algorithms work by using simple policies to sample control sequences, which are used to roll out the dynamics function and compute a sample-based approximation of the gradient of the objective function.
This approximate gradient is then used to update the sampling policy. 
Using complex dynamics and cost functions can make each rollout computationally expensive. To contend with this problem, one could use fewer samples to decrease computation, but this can increase the noise in the sample-based gradient, leading to poor performance.
%Another challenge for sampling-based MPC is that performance generally improves with an increasing number of samples, as it reduces the gradient noise.
%Performance generally improves with an increasing number of samples, as it reduces the gradient noise.
%However, this can be prohibitively expensive, especially for resource-constrained platforms under strict time constraints.
%Given the fact that we cannot compute the true gradient and wish to minimize the number of samples used, our objective is to take the components which would form the noisy gradient and learn how to more effectively update the control distribution.

In this paper, our objective is to  \emph{learn} how to more effectively update the control distribution with a small number of samples. To this end, we employ imitation learning to train fast, low-sample controllers to imitate an expert which makes use of additional samples. The learned optimizer is better able to integrate information in the sample-constrained regime. 
Our key contributions are that we: 
\begin{enumerate}
\item Leverage the gradient-based interpretation of many sampling-based MPC algorithms and show how to improve performance by learning a better update rule.
\item Propose to use structured sampling techniques to provide more information to the learned update than is contained in the noisy gradient, which enables us to make better use of fewer samples.
%\item Propose multiple novel neural network architecture and algorithm-specific decisions to improve performance of the learned optimizer.
\item Empirically evaluate our proposed approach on multiple simulated robotics tasks, in which the learned optimizer has restricted access to samples.
\end{enumerate}
Our experiments show that the learned controller is indeed able to make better use of fewer samples while remaining competitive or outperforming the expert with the same number of samples.
%Reducing the number of samples while achieving similar performance has the potential to improve the utility of sampling-based MPC controllers on embedded systems.
%This indicates the potential to improve the utility of sampling-based MPC controllers on embedded platforms.
This illustrates the utility of the learning-to-optimize framework in the domain of control and indicates the potential to improve the viability of sampling-based MPC controllers on embedded platforms.

\section{Background}
\subsection{Model Predictive Control}
We consider the problem of controlling a discrete-time stochastic dynamical system with states $x_t \in \real^N$ and controls $u_t \in \real^M$.
The system chooses controls using a policy $\pi_{\theta_t}$ with parameters $\theta_t \in \Theta$, where $\Theta$ is the set of feasible parameters.
After applying the control, the system incurs the instantaneous cost $c(x_t, u_t)$ and transitions to the next state $x_{t+1}$ according to the dynamics
\vspace{-1ex}
\begin{equation}
x_{t+1} \sim f(x_t, u_t),
\end{equation}
where $f:\real^N \times \real^M \rightarrow \real^N$ is a stochastic transition map. 
Over a time horizon $H$, we sample a control sequence $U_t \triangleq (u_t, u_{t+1}, \dots, u_{t+H-1})$, which results in a state trajectory $X_t \triangleq (x_t, x_{t+1}, \cdots, x_{t+H})$.
The total cost incurred is
\vspace{-0.5ex}
\begin{equation}
C(X_t, U_t) = \sum_{h=0}^{H-1} c(x_{t+h}, u_{t+h}) + c_{term}(x_{t+H}),
\vspace{-0.5ex}
\end{equation}
where $c_{term}(\cdot)$ is a terminal cost function.
The goal of MPC is to find the optimal set of parameters $\bm{\theta}_t \triangleq (\theta_t, \theta_{t+1}, \cdots, \theta_{t+H-1})$ for the sequence of policies $\bm{\pi}_{\bm{\theta}_t} \triangleq (\pi_{\theta_t}, \pi_{\theta_{t+1}}, \cdots, \pi_{\theta_{t+H-1}})$.
At each time step, we solve
\vspace{-0.5ex}
\begin{equation}
\vspace{-0.5ex}
\bm{\theta}_t \leftarrow \argmin_{\bm{\theta} \in \bm{\Theta}} J(\bm{\pi}_{\bm{\theta}}; x_t),
\label{eq:opt}
\end{equation}
where $J(\cdot)$ is a statistic defined on cost $C(X_t, U_t)$ such that its minimum occurs at the optimal $\bm{\theta}_t$.
%Usually we assume the policies $\bm{\pi}_{\bm{\theta}_t}$ are open-loop, but because we re-plan at each time step by solving \Cref{eq:opt}, we effectively have a state-feedback controller.
In general, we do not have access to the true dynamics function $f$ and instead approximate it with the model $\hat{f}$, corresponding to the surrogate statistic $\hat{J}(\bm{\pi}_{\bm{\theta}}; x_t)$.
%One popular choice of statistic is $\hat{J}(\bm{\pi}_{\bm{\theta}}; x_t) = \expect{\bm{\pi}_{\bm{\theta}}, \hat{f}}{C(X_t, U_t) | x_t}$, which estimates the expected $H$-step future costs using the approximate model.

This optimization problem can only be approximated in practice due to real-time constraints.
%, usually by an iterative algorithm like gradient descent.
One commonly applied heuristic is to bootstrap the previous approximate solution as an initialization for the current problem.
This is effective because the optimization problems between two consecutive time steps share all control variables except the first and last.
If our solution from the previous problem is $\bm{\theta}_{t-1}$, then our warm start for the current problem is given by
\begin{equation}
\tilde{\bm{\theta}}_t = \Phi(\bm{\theta}_{t-1}),
\label{eq:shift_operator}
\end{equation}
where $\Phi(\cdot)$ is called the shift operator~\cite{wagener2019online}.
A common choice is $\tilde{\bm{\theta}}_t = (\theta_{t+1}, \theta_{t+2}, \dots, \theta_{t+H-1}, \bar{\theta})$, where $\bar{\theta}$ is a new parameter which reflects the expected final action.
%This parameter is often set to a constant, such as zero, or repeats the last set of parameters along the horizon.

Recent work by Wagener et al. \cite{wagener2019online} showed that many common MPC algorithms fall under the framework of an online learning algorithm known as dynamic mirror descent (DMD) \cite{hall2013dynamical}.
Online learning involves interactions between a learner and an environment over $T$ rounds.
In our case, the learner is the MPC algorithm, which in round $t$ plays the decision $\tilde{\bm{\theta}}_t \in \bm{\Theta}$, the shifted policy parameter sequence, along with side information $u_{t-1}$, the control applied to the real system.
The per-round loss is defined as $\ell_t(\cdot) = \hat{J}(\cdot; x_t)$, which is selected by the environment via the state transition. % to state $x_t$ after applying control $u_{t-1}$.
At round $t$, DMD updates the parameters by the rule:
\begin{equation}
\bm{\theta}_t \leftarrow \argmin_{\bm{\theta} \in \bm{\Theta}} \inner{\gamma_t g_t}{\bm{\theta}} + D_{\psi}(\bm{\theta}||\tilde{\bm{\theta}}_t),%\quad
%\tilde{\bm{\theta}}_t = \Phi(\bm{\theta}_t),
\label{eq:dmd}
\end{equation}
where $g_t = \nabla \ell_t(\tilde{\bm{\theta}}_t)$, $\gamma_t > 0$ is the step size, %$\Phi$ is called the shift model, 
and $D_{\psi}(\bm{\theta}||\bm{\theta'}) = \psi(\bm{\theta}) - \psi(\bm{\theta}') - \inner{\nabla \psi(\bm{\theta}')}{\bm{\theta} - \bm{\theta}'}$ is the Bregman divergence generated by a strictly convex function $\psi$ on $\bm{\theta}$.
%This Bregman divergence acts as a regularizer to keep $\bm{\theta}$ close to $\tilde{\bm{\theta}}$.
%For MPC, the shift model is chosen to be the shift operator defined in \Cref{eq:shift_operator}.
Solving MPC with DMD is known as \textit{DMD-MPC} and includes a family of common MPC algorithms as special cases~\cite{wagener2019online}.

\subsection{Sampling-Based Model Predictive Control}
A popular, practical sampling-based MPC algorithm is Model Predictive Path Integral (MPPI) control \cite{williams2016aggressive, williams2017information}, which is a special case of DMD-MPC under certain  choices of objective function, control distribution, and Bregman divergence~\cite{wagener2019online}.
Specifically, we assume the policies are open loop and choose the exponential utility for the objective:
\begin{equation}
\ell_t(\bm{\theta}) = -\log \Expect{\bm{\pi}_{\bm{\theta}}, \hat{f}}{\exp \Big(-\frac{1}{\lambda} C(X_t, U_t) \Big)},
\end{equation}
where $\lambda > 0$ is a scaling parameter, also known as the temperature.
Since we generally assume that the cost function is non-differentiable with respect to $\bm{\theta}$, we instead compute the gradients via a likelihood-ratio derivative:
\begin{equation}
\small
\nabla \ell_t(\bm{\theta}) = - \frac{\Expect{\bm{\pi}_{\bm{\theta}}, \hat{f}}{\exp \Big(-\frac{1}{\lambda} C(X_t, U_t) \Big) \nabla_{\bm{\theta}}\log \pi_{\bm{\theta}}(U_t)}}{\Expect{\bm{\pi}_{\bm{\theta}}, \hat{f}}{\exp \Big(-\frac{1}{\lambda} C(X_t, U_t) \Big) }},
\end{equation}
We approximate these expectations with Monte Carlo sampling, which results in a convex combination of gradients:
\vspace{-1.5ex}
\begin{equation}
\nabla \ell_t(\bm{\theta}) = -\sum_{i=1}^N w_i \nabla_{\bm{\theta}} \log \pi_{\bm{\theta}}(U_t),
\label{eq:mc_gradients}
\vspace{-0.5ex}
\end{equation}
with weights $w_i$ defined by the softmax operation
\vspace{-1ex}
\begin{equation}
w_i = \frac{e^{ -\frac{1}{\lambda} C(X_t^{(i)}, U_t^{(i)}) }}{\sum_{j=1}^N e^{ -\frac{1}{\lambda} C(X_t^{(j)}, U_t^{(j)}) }}.
\vspace{-0.5ex}
\label{eq:sample_weights}
\end{equation}
Because of this estimation, the gradients will be noisy, with noise that scales inversely with number of samples.
The more samples we use, the more exact our approximations will be.

Next, we choose the Bregman divergence to be the KL divergence and the distribution to be a factorized Gaussian:
\begin{equation}
\small
\pi_{\bm{\theta}}(U_t) = \prod_{h=0}^{H-1} \pi_{\theta_h}(u_{t+h}) 
= \prod_{h=0}^{H-1} \mathcal{N}(u_{t+h}; \mu_{t+h}, \Sigma_{t+h}), 
\end{equation}
for some mean vectors $\mu_{t+h}$ and covariance matrices $\Sigma_{t+h}$.
Under these assumptions, Wagener et al. \cite{wagener2019online} showed that the solution to \Cref{eq:dmd} becomes
\vspace{-0.5ex}
\begin{equation}
\small
\begin{aligned}
\mu_{t+h} &= (1-\gamma_t^\mu) \tilde{\mu}_{t+h} + \gamma_t^\mu \frac{\Expect{\bm{\pi}_{\bm{\theta}}, \hat{f}}{e^{ -\frac{1}{\lambda} C(X_t, U_t) } u_{t+h}}}{\Expect{\bm{\pi}_{\bm{\theta}}, \hat{f}}{e^{ -\frac{1}{\lambda} C(X_t, U_t)} }} \\
\Sigma_{t+h} &= (1-\gamma_t^\sigma) \tilde{\Sigma}_{t+h} + \gamma_t^\sigma \frac{\Expect{\bm{\pi}_{\bm{\theta}}, \hat{f}}{e^{ -\frac{1}{\lambda} C(X_t, U_t) } m_{t+h}m_{t+h}^T}}{\Expect{\bm{\pi}_{\bm{\theta}}, \hat{f}}{e^{ -\frac{1}{\lambda} C(X_t, U_t)} }},
\end{aligned}
\label{eq:MPPI_update_exact}
\end{equation}
where $m_{t+h} = u_{t+h} - \mu_{t+h}$ and $\gamma_t^\mu$ and $\gamma_t^\sigma$ are step sizes for the mean and covariance, respectively.
When we use the Monte-Carlo approximations to the gradients in \Cref{eq:mc_gradients}, this results in the following update:
\vspace{-0.5ex}
\begin{equation}
\begin{aligned}
\mu_{t+h} &= (1-\gamma_t^\mu) \tilde{\mu}_{t+h} + \gamma_t^\mu \sum_{i=1}^N w_i u_{t+h}^{(i)} \\
\Sigma_{t+h} &= (1-\gamma_t^\sigma) \tilde{\Sigma}_{t+h} + \gamma_t^\sigma \sum_{i=1}^N w_i m_{t+h}^{(i)}m_{t+h}^{{(i)}^T},
\end{aligned}
\label{eq:MPPI_update}
\vspace{-0.5ex}
\end{equation}
This update equation is the same as the MPPI update rule when $\gamma_t^\mu = 1$ and we do not update the covariance. 

%Since we update the covariance matrix each time step, we have to choose a factorization which is flexible yet tractable.
%Standard implementations of MPPI choose $\Sigma_{t+h} = \sigma_u I_{M\times M}$, where $\sigma_u$ is a scalar and $I_{M\times M}$ is the $M$-dimensional identity matrix.
%While this parameterization is simple, it forces the covariance to be the same across all control dimensions. 
%This choice may be suboptimal for many applications.
%Instead, as in Bhardwaj et al. \cite{bhardwaj2021fast}, we choose to assign different covariances to each dimension of the controls in the form $\Sigma_{t+h} = \diag{\sigma_u}$, where $\sigma_u = [\sigma_1, \dots, \sigma_M]$.
%Each term $\sigma_i$ is then adapted according to \Cref{eq:MPPI_update}.

\subsection{Learning to Optimize Framework}
Rather than hand design an update rule tailored to a specific subclass of problems, the learning-to-optimize approach aims to learn a sequential update rule from experience.
For a set of optimizee parameters $\theta \in \Theta$ and objective function $\ell(\theta)$, we find the minimizer $\theta^* = \argmin_{\theta \in \Theta} \ell(\theta)$ with an iterative algorithm that has the update rule
\vspace{-0.5ex}
\begin{equation}
\theta_{t+1} = m_{\phi}(\theta_t, t),
\vspace{-0.5ex}
\label{eq:l2o}
\end{equation}
where $m$ is the learned optimizer, which can be of any parameterized function class with parameters $\phi$.

The majority of approaches to learning-to-optimize differentiate through the optimization process using gradient descent \cite{andrychowicz2016learning, ravi2016optimization, lv2017learning, chen2017learning, chen2020training} or use reinforcement learning \cite{li2016learning, wang2016learning, li2017learning, bello2017neural}.
% with the goal of improving the training process of neural networks.
However, we do not assume that the optimization process is end-to-end differentiable and wish to avoid the high sample complexity that often hinders reinforcement learning.
Instead, we opt to use imitation learning to train the optimizer.
Chen et al. \cite{chen2020training} also make use of imitation learning, in which the experts are common hand-designed optimizers that have access to full gradient information.
%An imitation loss is combined with differentiating through the optimization process in order to improve generalization of the learned optimizer.
%However, our use of imitation learning is slightly different, in that our learned optimizers have access to less information (samples) than the expert demonstrator.
However, in our case, the learned optimizers only have access to noisier gradients than the expert demonstrator and therefore less information.

Another major difference from prior work is that most literature in this area targets optimizing deep neural networks.
As such, they must contend with the large parameter space of these models.
Andrychowicz et al.~\cite{andrychowicz2016learning} proposed to use a coordinate-wise optimizer, in which the parameters of the optimizer are shared across updates for all optimizee parameters.
A downside to this approach is that it throws away potentially useful information for improving the learned update.
%When the optimizer is implemented with a recurrent network, differences in the hidden state result in varying behaviors for each coordinate.
%With a moderate number of parameters, we can jointly optimize them to capture more complex relationships.
Instead, since we have a moderate number of parameters, we can jointly optimize the entire planning horizon of control distribution parameters in order to capture relationships between time steps.

%Another interesting direction is to generate a mathematical update rule rather than represent it with a black-box function approximator \cite{bello2017neural}.
%However, we are ultimately more interested in performance than interpretability of the learned optimizer.
%Therefore, we use neural networks to represent the learned optimizer in all experiments.
%To our knowledge, we are the first to bring the learning to optimize framework to solving problems in the domain of control.

\section{Learning to Optimize for Control}
\subsection{Design of the Learnable Optimzier}
\label{sec:architecture}

As shown in the previous section, the MPPI update rule in \Cref{eq:MPPI_update} corresponds to performing mirror descent with an approximate gradient computed from $N$ samples. Fewer samples results in a worse approximation and, therefore, a noisier update. 
%One possible avenue for improving performance would be to employ more advanced first-order methods, such as momentum \cite{nesterov27method, qian1999momentum} or scaling updates based on the gradient history \cite{duchi2011adaptive, zeiler2012adadelta, kingma2014adam, reddi2019convergence}.
One possible avenue for improving performance would be to employ more advanced first-order methods \cite{nesterov27method, qian1999momentum, duchi2011adaptive, zeiler2012adadelta, kingma2014adam, reddi2019convergence}.
%Instead, we follow the learning-to-optimize framework and propose to replace the update rule with a \emph{learned} optimizer.
However, by adopting the learning-to-optimize framework and replacing the update rule with a learned optimizer, we can potentially do better than a manually specified update.
A naive approach would be to follow \Cref{eq:l2o} and use the noisy gradient as input to the learned optimizer to produce the updated parameters.
%This may be sufficient if our goal was to improve convergence speed to a local optima.
However, our objective is to learn how to mitigate the effect of a low number of samples on gradient noise, and the computation of the noisy gradient itself potentially throws away information that may be useful for improving the update.
For instance, looking at \Cref{eq:MPPI_update}, we are simply computing a weighted sum of the samples.
This collapses the information in each trajectory sample and its corresponding cost into a single vector.
Therefore, we propose instead to use the individual components which form the gradient directly.

% One potential choice for learned updates follows from rewriting \Cref{eq:MPPI_update} in a more general form:
% \begin{equation}
% \mu_{t+h}, \Sigma_{t+h} = m_{\phi}\Big(\tilde{\mu}_{t+h}, \tilde{\Sigma}_{t+h}, w^{(1:N)}_t, u_{t+h}^{(1:N)}\Big),
% \end{equation}
% where $w^{(1:N)}_t$ are the sample weights, $u_{t+h}^{(1:N)}$ and are all control samples for time step $t+h$, and $m$ is a function with a set of parameters $\phi$.
% In the standard MPPI update, the parameters could be the step sizes $\gamma_t^\mu$ and $\gamma_t^\sigma$.
% However, more generally, this update could be any black box function, and we can learn its parameters.

From \Cref{eq:MPPI_update}, we can see that the update is a function of the current mean $\tilde{\mu}_{t+h}$ and covariance $\tilde{\Sigma}_{t+h}$, sample weights $w^{(1:N)}_t$, and control samples $u_{t+h}^{(1:N)}$.
The sample weights themselves are actually a function of the total trajectory costs $C^{(1:N)}_t$, where $C_t = C(X_t, U_t)$.
One potential choice would be to make each of these terms an input to the learned update:
\vspace{-0.5ex}
\begin{equation}
\mu_{t+h}, \Sigma_{t+h} = m_{\phi}\Big(\tilde{\mu}_{t+h}, \tilde{\Sigma}_{t+h}, C^{(1:N)}_t, u_{t+h}^{(1:N)}\Big).
\end{equation}
A limitation of this choice of parameterization is that it assumes independence of the updates between time steps in the rollouts. 
While this is the case for vanilla MPPI~\cite{williams2016aggressive, williams2017information}, we could potentially learn a better update by incorporating information across time steps.
% As such, we propose to learn a joint update
% \begin{equation}
% \bm{\mu}_t, \bm{\Sigma}_t = m_{\phi}\Big(\tilde{\bm{\mu}}_t, \tilde{\bm{\Sigma}}_t, w^{(1:N)}_t, U_t^{(1:N)}\Big)
% \label{eq:learned_update},
% \end{equation}
% where $\bm{\mu}_t \triangleq (\mu_{t}, \mu_{t+1}, \dots, \mu_{t+H-1})$, $\bm{\Sigma}_t \triangleq (\Sigma_{t}, \Sigma_{t+1}, \dots, \Sigma_{t+H-1})$, and $\tilde{\bm{\mu}}_t$ and $\tilde{\bm{\Sigma}}_t$ are defined similarly.
% However, the direct use of all control samples would result in a large input space for the function approximator.
% If we use a fully-connected or recurrent neural network architecture, this will result in a large number of parameters to learn, making optimization difficult.
% The simplest way to address this issue would be to instead compute the MPPI update first and then feed it to the network, along with the current mean and covariance.
% However, by doing so, we may potentially throw away useful information for learning a better update.
However, if we parameterize the optimizer with a fully-connected or recurrent neural network architecture, this would result in a large number of parameters to learn, making optimization difficult.
Instead, we alter the way in which we sample from the Gaussian policies to remedy this explosion in the dimensionality.% of the input space and parameter count.

As proposed by Bhardwaj et al. \cite{bhardwaj2021fast}, we make use of low-discrepancy Halton sequences \cite{halton1964algorithm} to generate samples from the Gaussian policies.
Normal pseudo-random sequences often result in clusters of sampled points, leaving many regions of the parameter space untouched.
Low-discrepancy sequences are a deterministic alternative that alleviate this problem by correlating each point.
%Moreover, they have a faster rate of convergence for estimating moments of distributions \cite{niederreiter1992random}.
A $D$-dimensional Halton sequence $x_1, x_2, \dots, x_N$, in which $x_i \in \real^D$ is generated by
\vspace{-0.5ex}
\begin{equation}
x_i = (\phi_{p_1}(i), \dots, \phi_{p_D}(i)),\
\phi_{p_b}(i) = \sum_{j=1}^\infty a_j(p_b) p_b^{-j},
\vspace{-0.5ex}
\end{equation}
where $p_1, \dots, p_D$ are consecutive prime numbers and $a_j(p_b) \in \{0, 1, \dots, p_b-1 \}$ such that the condition $i = \sum_{j=1}^\infty a_j(p_b) p_b^{j-1}$ holds.
The Halton sequence is sampled once at the beginning of the rollout and then transformed using the mean and covariance of the Gaussian policy.
%Therefore, all sampled control sequences are a deterministic function of the current mean and covariance and can be excluded from \Cref{eq:learned_update} without loss of information, assuming the function approximator is expressive enough.
While this can improve the performance of sampling-based MPC, the main benefit is that it makes all sampled control sequences a deterministic function of the current mean and covariance. 
Therefore, the sampled control sequences can be excluded from the learned update without loss of information.

%Rather than use the noisy gradient as input to our optimizer and optimize each time step independently, we leverage the structured nature of these samples to learn an update that makes direct use of \emph{all} of the cost information across the entire trajectory.
As such, rather than optimizing each time step independently, we leverage the structured nature of these samples to learn an update that optimizes the entire trajectory jointly using only cost information.
The resulting update is then
\begin{equation}
\bm{\mu}_t, \bm{\Sigma}_t = m_{\phi}\Big(\tilde{\bm{\mu}}_t, \tilde{\bm{\Sigma}}_t, C^{(1:N)}_t\Big)
\label{eq:learned_update},
\end{equation}
where $\bm{\mu}_t \triangleq (\mu_{t}, \mu_{t+1}, \dots, \mu_{t+H-1})$, $\bm{\Sigma}_t \triangleq (\Sigma_{t}, \Sigma_{t+1}, \dots, \Sigma_{t+H-1})$, and $\tilde{\bm{\mu}}_t$ and $\tilde{\bm{\Sigma}}_t$ are defined similarly.
We can think about the Halton sequence as giving us a sense of what the environment and cost landscape is like around the current state.
Since the learned optimizer can potentially make better use of its inputs than the expert, we may be able to more effectively use fewer samples while maintaining similar performance.
% Another modification we make is to replace the sample weights $w^{(1:N)}_t$ and directly use the total trajectory costs $C(X_t, U_t)$.
% As we train our learned updates via imitation learning (see \Cref{sec:training}), we empirically found that policies learned with access to fewer samples than the expert performed better with this parameterization.
% \jake{One potential explanation for this behavior is that when using the sample weights directly, the true underlying MPPI update of the expert is a fairly simple function (\Cref{eq:MPPI_update}).
% During training, the network appears to learn a function which behaves similarly to this update, even when given access to fewer samples.
% As such, it fails to make better use of the samples and cannot outperform MPPI with access to the same number of samples.}

Finally, we note that \Cref{eq:MPPI_update} is a convex combination of the previous control parameters and the weighted samples.
We can actually think about MPC as a form of recurrent network, with the warm-started control distribution as our form of memory about the previous time steps.
From this perspective, the step size is acting as a gating term which modulates how much information we preserve about our history.
The hidden state update in a gated recurrent unit (GRU) \cite{cho2014learning} is of the same form, except the multiplicative gating term is also learned.
Inspired by this similarity, we use the following update:
\begin{equation}
\begin{gathered}
g_t^\mu, g_t^\sigma, h_t^\mu, h_t^\sigma = m_{\phi}(\tilde{\bm{\mu}}_t, \tilde{\bm{\Sigma}}_t, C_t^{(1:N)}) \\
\bm{\mu}_t = (1 - g_t^\mu) \odot \tilde{\bm{\mu}}_t + g_t^\mu \odot h_t^\mu\\
\bm{\Sigma}_t = (1 - g_t^\sigma) \odot \tilde{\bm{\Sigma}}_t + g_t^\sigma \odot h_t^\sigma,
\end{gathered}
\label{eq:learned_update_final}
\end{equation}
where $\odot$ is the Hadamard product. 
Here, $g_t^\mu, g_t^\sigma$ are the learned gating terms, which are passed through a sigmoid to ensure they are between zero and one.
Meanwhile, $h_t^\mu, h_t^\sigma$ can be interpreted as the updates proposed by the network.
In our experiments, this choice of parameterization significantly outperformed a simple fully-connected network.

\subsection{Imitation Learning for Training the Optimizer}
\label{sec:training}

\begin{algorithm}[t!]
\SetAlgoLined
\KwIn{Initial bootstrapped dataset $\mathcal{D}$, initial policy $\bm{\pi}_{{\tilde{\bm{\theta}}}_1}$, initial state distribution $\rho$}
\KwParam{Iterations $K$, probabilities $\{\beta_k\}_{k=1}^K$, rollouts per iteration $R$}
\For{$k = 1, 2, \dots, K$}{
    Initialize dataset $\mathcal{D}_i \leftarrow \emptyset$ \\
    \For{r = 1, 2, \dots, R}{
        Sample initial state $x_1 \sim \rho$ \\
         $\tilde{\bm{\theta}}_{1:T}$, $C_{1:T}^{(1:M)}$, $\bm{\theta}_{1:T}^{expert}$ $\leftarrow$ Rollout($x_1, \bm{\pi}_{\tilde{\bm{\theta}}_1}, \beta_k$)\\
        Append $\mathcal{D}_i \leftarrow \mathcal{D}_i \cup \{( \tilde{\bm{\theta}}_{1:T}, C_{1:T}^{(1:M)}, \bm{\theta}_{1:T}^{expert})\}$
    }
    Aggregate datasets $\mathcal{D} \leftarrow \mathcal{D} \cup \mathcal{D}_i$ \\
    Train optimizer parameters $\phi$ on $\mathcal{D}$
}
\caption{\textsc{DAgger} Training Loop}
\label{alg:DAgger}
\end{algorithm}

Unlike prior work in learning-to-optimize, we cannot assume that the optimization process itself is differentiable, as it occurs online via interactions with the environment.
Even in the simulated case, we do not want to assume that everything has been implemented in a differentiable fashion.
We could use reinforcement learning (RL), although it generally has high sample complexity and may be slow to learn.
Since we have access to a tuned optimal controller, imitation learning is a promising direction for training the optimizer.
Our expert is an MPPI controller with unrestricted access to samples. 
That is, we provide the controller with as many samples as needed to achieve good performance.
The learner is also an MPPI controller, but it has access to fewer samples, and the standard update is replaced with the learned optimizer.
%The goal is then to learn an update which enables the learner to remain competitive in performance with the expert by making better use of fewer samples.
%Note that the expert is only run for a single iteration.
%However, we could perform multiple iterations of the MPPI update to potentially get a better target control distribution.
%The learner is also run for a single iteration, as we assume that the learned optimizer can perform a sufficient update in a single iteration, regardless of the number of iterations the expert was run.
In our preliminary experiments, we tried using standard behavioral cloning, in which we collect a dataset of expert demonstrations and train a policy offline via regression.
However, this did not work well due to covariate shift between the expert and learner distributions.
Instead, we used \textsc{DAgger} \cite{ross2011reduction} to perform imitation learning, which is an interactive algorithm that aims to combat issues of covariate shift.
The algorithm queries an expert online for corrective labels on learner visited states.
We outline the main loop of \textsc{DAgger} in Algorithm \ref{alg:DAgger}.

\begin{algorithm}[t!]
\SetAlgoLined
\KwIn{State $x_1$, policy $\bm{\pi}_{\tilde{\bm{\theta}}_1}$, probability $\beta_k$}
\KwParam{Rollout length $T$, expert samples $N$, learner samples $M$}
\KwOut{Shifted parameters $\tilde{\bm{\theta}}_{1:T}$, sample costs $C_{1:T}^{(1:M)}$, expert decisions $\bm{\theta}_{1:T}^{expert}$}
\For{$t = 1, 2, \dots, T$}{
    Sample controls from policy $\{U_t^{(i)}\}_{i=1}^N \sim \bm{\bm{\pi}}_{\tilde{\bm{\theta}}_t}$ \\
    Sample $X_t^{(i)}$ from dynamics $\hat{f}$ using $U_t^{(i)}, x_t$ \\
    Compute costs $C_t^{(i)} \leftarrow C(X_t^{(i)}, U_t^{(i)})$ \\
    Compute sample weights with \Cref{eq:sample_weights} \\
    Update $\bm{\tilde{\theta}}_{t}$ to $\bm{\theta}_{t}^{expert}$ using \Cref{eq:MPPI_update} \\
    Sample $b \sim U(0, 1)$ \\
    \eIf{$b \leq \beta_k$}{
        Set $\bm{\theta}_t \leftarrow \bm{\theta}_t^{expert}$
    }{
        Update $\bm{\tilde{\theta}}_{t}$ to $\bm{\theta}_{t}^{learn}$ using     \Cref{eq:learned_update_final} \\
        Set $\bm{\theta}_t \leftarrow \bm{\theta}_t^{learn}$
    }
Sample $u_t \sim \pi_{\theta_{t}}$ or use mean $u_t \leftarrow \mu_t$ \\
Apply control to system $x_{t+1} \sim f(x_t, u_t)$ \\
Shift parameters $\tilde{\bm{\theta}}_{t+1} = \Phi(\bm{\theta}_{t})$
}
\caption{\textsc{DAgger} Rollout Function}
\label{alg:rollout}
\end{algorithm}

First, we begin by collecting a bootstrap dataset in which only the expert is run.
Next, each iteration $k$ of \textsc{DAgger}, we run $R$ rollouts according to Algorithm \ref{alg:rollout} by sampling some initial state $x_1$ from a known initial state distribution $\rho$.
During a rollout, at each time step, we apply the controls from the expert with probability $\beta_k$ and the learner with probability $1-\beta_k$.
The expert is always run in order to provide a corrective target for training the policy at the next iteration.
Both the expert and learner controllers use the same trajectory samples, although the learner only receives a subset of them.
%Only the form of the update applied to the policy distribution is different.
Generally, the mixing probabilities $\beta_k$ are set according to a schedule such that we run the learner more often in later iterations.
In our experiments, we set $\beta_k = p^{k}$ for some $p \in (0, 1)$.
After running the rollouts, we collect the warm-started control distribution parameters $\tilde{\bm{\theta}}_{1:T}$, the trajectory costs $C_{1:T}^{(1:M)}$, and the updated expert control distribution parameters $\bm{\theta}_{1:T}^{expert}$ into a dataset $\mathcal{D}_i$.
%While we compute $N$ samples for the expert, we only collect the $M \leq N$ used by the learner.
We only collect the $M \leq N$ samples used by the learner.
This data is then aggregated into our main dataset $\mathcal{D}$ to train the optimizer.

\section{Experiments}
\label{sec:experiments}

\textbf{Implementation Details.}
In all experiments, our MPPI implementation is a modified version of the one developed by Bhardwaj et al. \cite{bhardwaj2021fast}.
This implementation uses Halton sequences for generating control sequence samples and smooths the sampled trajectories with 3rd degree B-splines.
We use a fixed diagonal covariance for the sampling distribution and do not perform covariance adaptation.
All hyperparameters were tuned using a grid search, and the optimal number of samples is what the expert controller has access to during data generation and training.
Now, the optimal choice of hyperparameters may be different for a given number of samples.
Therefore, for a fair comparison, we tune the MPPI hyperparameters separately for each sample count used in our evaluation.
%This implementation additionally performs covariance adaptation, uses Halton sequences for generating samples, and smooths the sampled trajectories with B-splines of degree 3.
%As such, in addition to the standard MPPI parameters, we define separate step sizes for the mean and covariance and number of spline knots.
%All parameters were tuned using a grid search, and the optimal number of samples is what the expert controller has access to in our experiments.
%We report all of the parameters for each experiment in \Cref{tab:mppi_param}.
Both MPPI and the neural networks are implemented in PyTorch \cite{paszke2019pytorch}.

\textbf{Task Details.}
We evaluate on simulated tasks:
\begin{enumerate}
\item \textsc{Cartpole}: The task is to slide a cart along a rail to swing up the pole attached via an unactuated joint using only actuation from the cart. Both the expert and learner are given access to the true analytical dynamics. The initial position of the cart and pole are randomized at every episode, which lasts 200 time steps. An episode is considered successful if the pole is swung up with a linear and angular velocity near zero.
\item \textsc{Franka Reacher}: A 7 degree-of-freedom (DOF) Franka Panda robot arm must reach a target goal from a fixed starting pose. The goal is randomly selected at the beginning of each episode. Both the expert and learner use the same kinematic model described in Bhardwaj et al. \cite{bhardwaj2021fast}, which is different from the true dynamics of the simulator (Nvidia's Isaac Gym \cite{makoviychuk2021isaac}). Each episode lasts for 500 time steps and is considered successful if the end effector reaches the target position.
\item \textsc{Franka Obstacles}: This task is identical to \textsc{Franka Reacher}, except now there are two spherical obstacles placed in the environment which the arm must avoid. The obstacle and the goal positions are randomized at the beginning of each episode, which lasts for 600 time steps. An episode is considered successful if the end effector reaches the goal while avoiding collisions. 
\end{enumerate}

\textbf{Evaluation.}
We evaluate the performance of the learned optimizer by varying the number of samples, up to the amount used by the expert.
For each sample amount, we compare against a standard MPPI implementation with access to the same number of samples as the learned controller over 30 test rollouts.
All test rollouts use a fixed set of start states, goals, and obstacle locations.
This is achieved by setting the random seed value to a pre-defined test seed.
Our primary metric for comparison is success rate, which is defined as the percentage of times the task goal was achieved out of all trials.
In the \textsc{Franka Obstacles} task, the placement of obstacles is randomized according to a pre-specified distribution.
Therefore, this task allows us to evaluate the generalization capability of the learned optimizer to new environments which are drawn from a similar distribution.
Additionally, we report statistics of the end effector test trajectories.
Specifically, we compute a relative trajectory length and average jerk as the ratio between the statistics for the learned optimizer and baseline MPPI controller.
The trajectory length is averaged over all test runs, while the jerk is averaged over only successful test runs.

% \textbf{Baselines and Ablations.} 
% For each sample amount, we compare against a standard MPPI implementation with access to the same number of samples as the learned controller.
% Additionally, we perform ablations for different components of our network, including: 
% \begin{enumerate}
% \item Using a fully-connected network vs. our proposed gating architecture described in \Cref{sec:architecture}.
% \item Using sample weights from the softmax (\Cref{eq:sample_weights}) as input to the network vs. the total trajectory costs.
% \item Simultaneously optimizing the full horizon jointly with a single network vs. optimizing each time step independently.
% \end{enumerate}
% For the ablation which optimizes each time step independently, we apply the same network on each control per time step, similar to the coordinate-wise optimizer by Andrychowicz et al. \cite{andrychowicz2016learning}.
% In their case, the network is applied to each scalar parameter separately.
% Instead, we apply the network to all control parameters for a given time step.
% While the parameters are shared for the network applied to each time step, different behavior is encoded via separate network activations.

\textbf{Training Details.}
Prior work in learning-to-optimize made use of recurrent architectures which can account for the history of the gradients and optimization process.
While we could potentially benefit from such an architecture, we found that a simple multi-layer perceptron (MLP) was sufficient to learn powerful optimizers.
As such, in all experiments, the learned optimizer is represented with a two-layer MLP using \relu\ activation functions.
We use 1024, 2048, and 4096 hidden units per layer for the \textsc{Cartpole}, \textsc{Franka Reacher}, and \textsc{Franka Obstacles} tasks, respectively.
To prevent overfitting, all networks are regularized with dropout \cite{srivastava2014dropout} using a dropout probability of 0.1.
We use the \textsc{Adam} optimizer \cite{kingma2014adam} with a learning rate of $10^{-3}$ for \textsc{Cartpole} and \textsc{Franka Reacher} and $10^{-4}$ for \textsc{Franka Obstacles}.
%For networks that used the total trajectory costs, we normalize the costs based on the mean and standard deviation of the training dataset to speed up convergence.
We normalize the total trajectory costs based on the mean and standard deviation of the training dataset.
For all tasks, we bootstrapped the dataset with 1024 trajectories, in which only the expert's action was applied to the system.
We ran \textsc{DAgger} for 20 iterations with 128 rollouts per iteration and a mixing probability schedule $\beta_k = 0.8^k$.
For each iteration, we train the networks on the aggregated dataset for 1000 epochs with a batch size of 8.
%The dataset is divided into training and validation splits, where new trajectories collected with \textsc{DAgger} are appended only as training data and the validation set is held constant.
%After each epoch of training, the network is evaluated on held out validation data.
The best performing network evaluated on a held-out validation set is saved and used in the next iteration of \textsc{DAgger}.

\begin{table}[t]
\centering
\caption{Success Rate Across All Tasks.}
\begin{tabular}{c|c|cc}
\hrulethick
& \textbf{\# Samples} & \textbf{MPPI} & \textbf{L2O-MPC} \\
\hline
\multirow{3}{*}{\rotatebox[origin=c]{90}{\scriptsize\textsc{Cartpole}}} 
& 8 & 100.0  & 100.0 \\
& 4   & 60.00  & 96.67 \\ 
& 2   & 10.00  & 90.00 \\
\hline
\multirow{6}{*}{\rotatebox[origin=c]{90}{\scriptsize\textsc{Franka Reacher}}} 
& 64  & 100.0 & 100.0 \\
& 32  & 90.00 & 100.0  \\
& 16  & 63.33 & 100.0 \\
& 8   & 20.00 & 80.00 \\
& 4   & 10.00 & 63.33 \\ 
& 2   & 3.333  & 16.67 \\
\hline
\multirow{6}{*}{\rotatebox[origin=c]{90}{\scriptsize\textsc{Franka Obstacles}}} 
& 512 & 80.00 & 80.00 \\
& 256 & 76.67 & 80.00 \\
& 128 & 73.33 & 76.67 \\
& 64  & 63.33 & 76.67 \\
& 32  & 33.33 & 66.67 \\
& 16  & 6.667 & 46.67 \\
\hrulethick
\end{tabular}
\label{tab:results}
\vspace{-2ex}
\end{table}

\section{Results}
We report the success rate for all tasks in \Cref{tab:results} and refer to standard MPPI by \textbf{MPPI} and MPC with the learned optimizer by \textbf{L2O-MPC}.
Success rate is computed for each task based on the criteria discussed in \Cref{sec:experiments}.
For \textsc{Franka Obstacles}, not every randomly generated scenario is feasible.
Hence, there is an upper-bound of an 80\% success rate for both \textbf{MPPI} and \textbf{L2O-MPC}.
We can see that the performance of \textbf{MPPI} quickly drops off as the number of control sequence samples is reduced. 
For \textsc{Cartpole}, the performance of \textbf{L2O-MPC} remains fairly consistent even with a lower number of samples.
While the performance drop is more pronounced in the Franka experiments, \textbf{L2O-MPC} still consistently matches or outperforms \textbf{MPPI} at each sample amount.
In \textsc{Franka Reacher}, \textbf{L2O-MPC} is able to withstand a $4 \times$ decrease in the number of samples while still achieving an $100\%$ success rate.
Similarly, in \textsc{Franka Obstacles}, \textbf{L2O-MPC} only incurs a $4\%$ decrease in performance under an $8\times$ decrease in number of samples.
This illustrates that \textbf{L2O-MPC} is successfully able to generalize to new environments similar to those on which it was trained.

\begin{figure}[t]
\centering
\hspace{-1ex}
\includegraphics[width=0.48\textwidth]{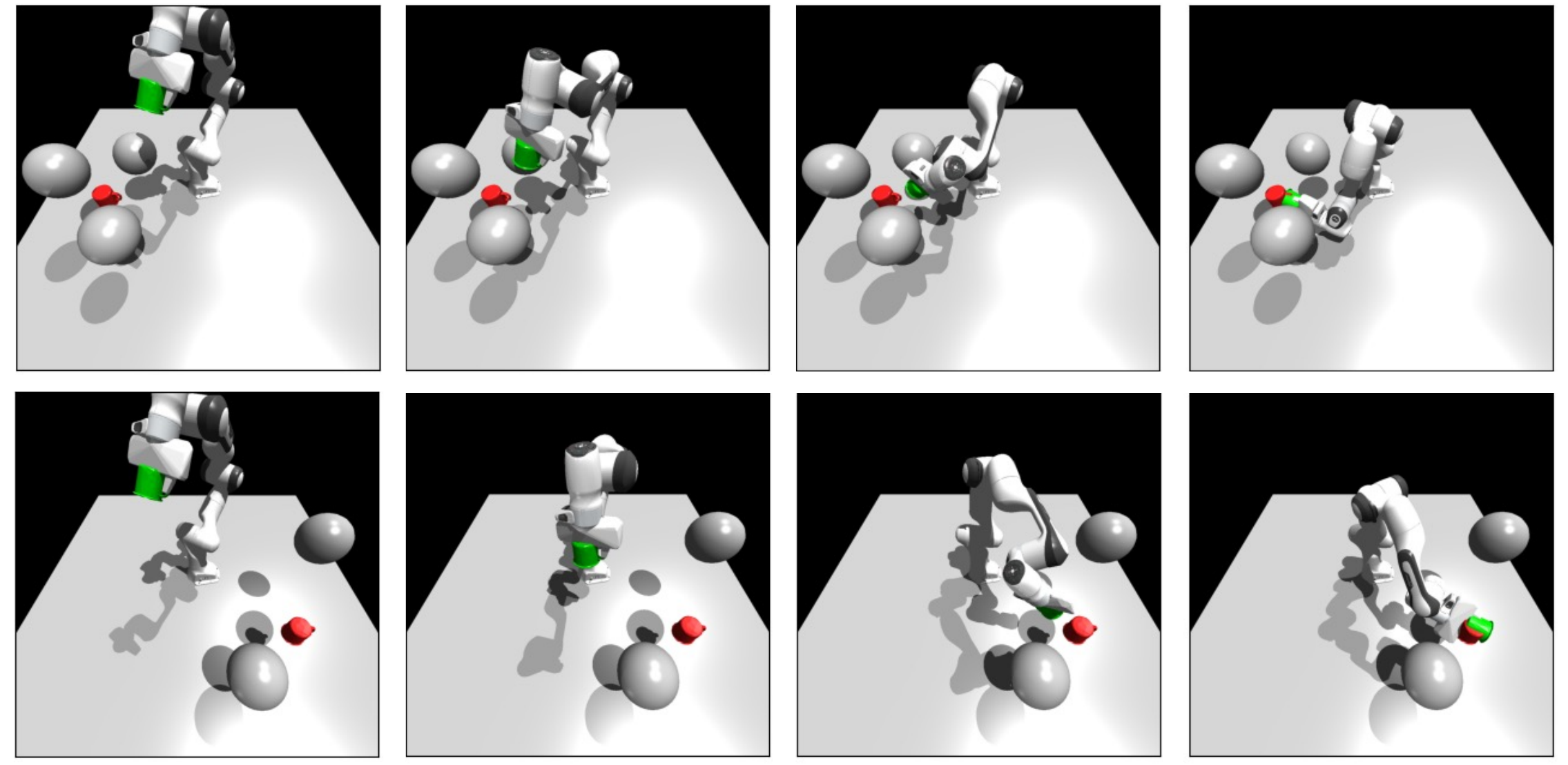}
\caption{Example trajectories of the \textsc{Franka Obstacles} task, in which the Franka arm end effector (green) is tasked with reaching the goal (red) while avoiding obstacles. The top row is a novel environment with three obstacles, while the bottom row is an environment from the test set.}
\label{fig:images}
\end{figure}

\begin{figure}[t]
\centerline{
\hspace{-2ex}
\begin{subfigure}[b]{0.19\textwidth}
    \centering
    \includegraphics[width=\textwidth]{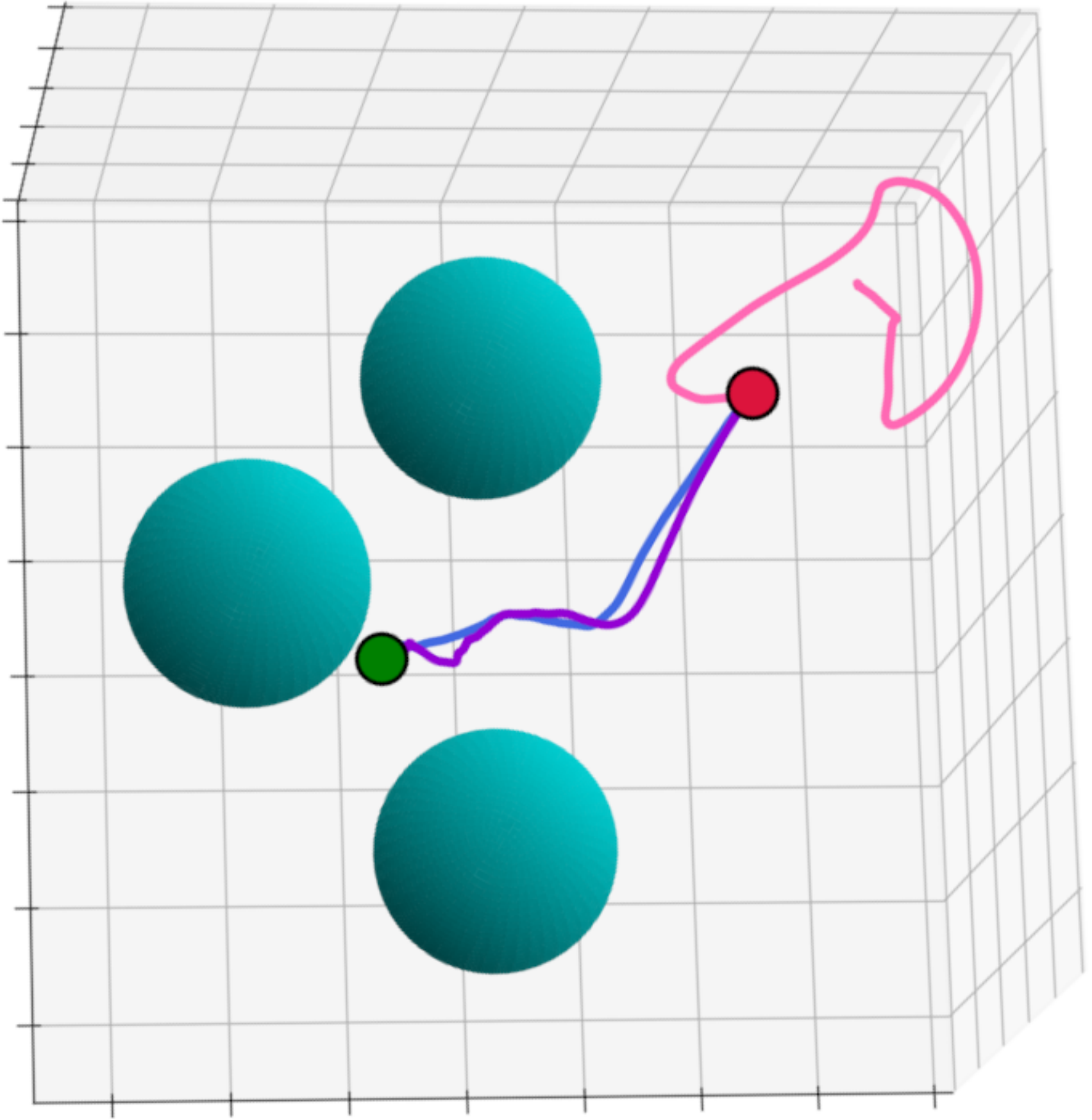}
\end{subfigure}
\hspace{1.5ex}
\begin{subfigure}[b]{0.23\textwidth}
    \centering
    \includegraphics[width=\textwidth]{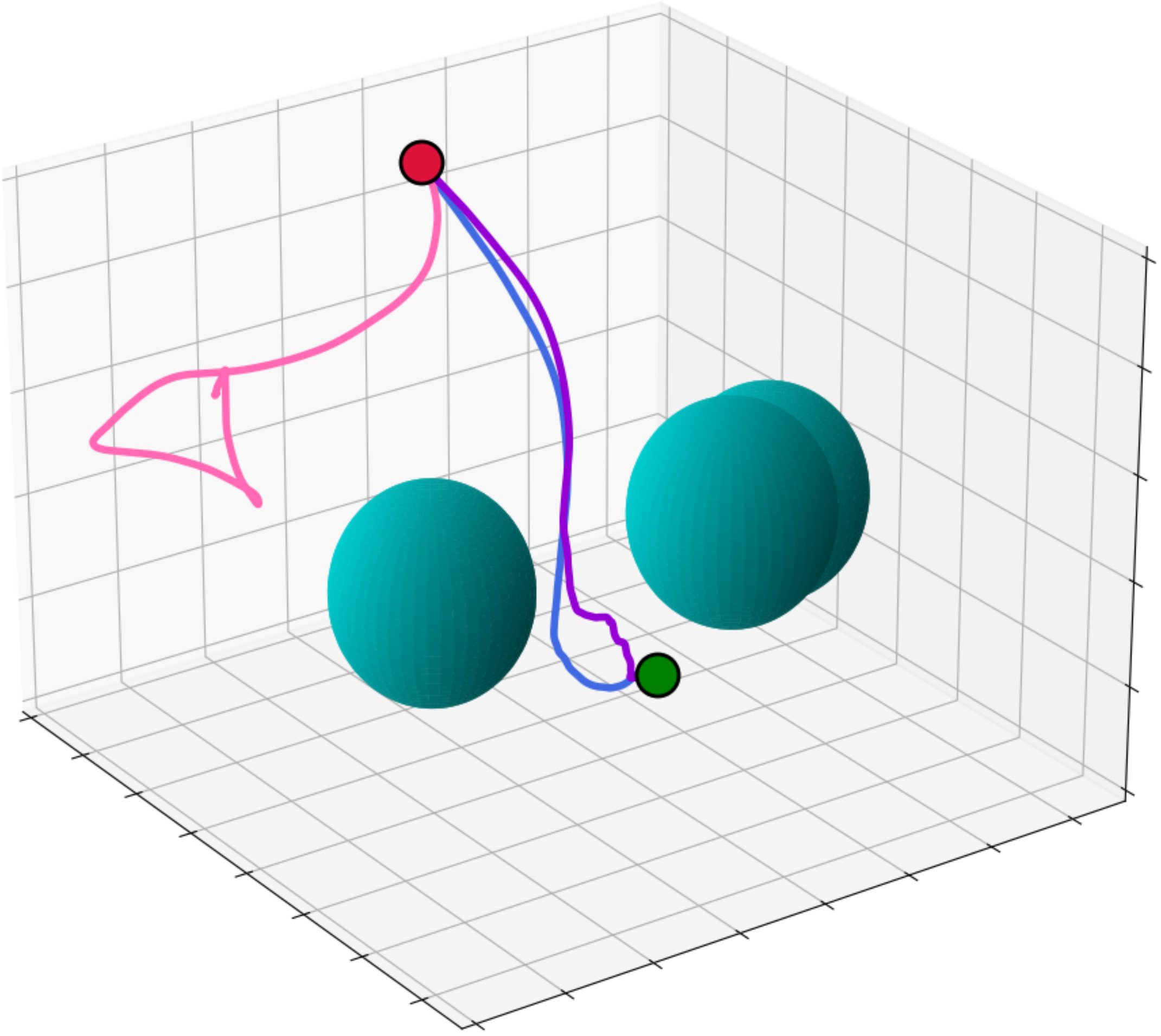}
\end{subfigure}
}
\vspace{2ex}
\centerline{
\begin{subfigure}[b]{0.185\textwidth}
    \centering
    \includegraphics[width=\textwidth]{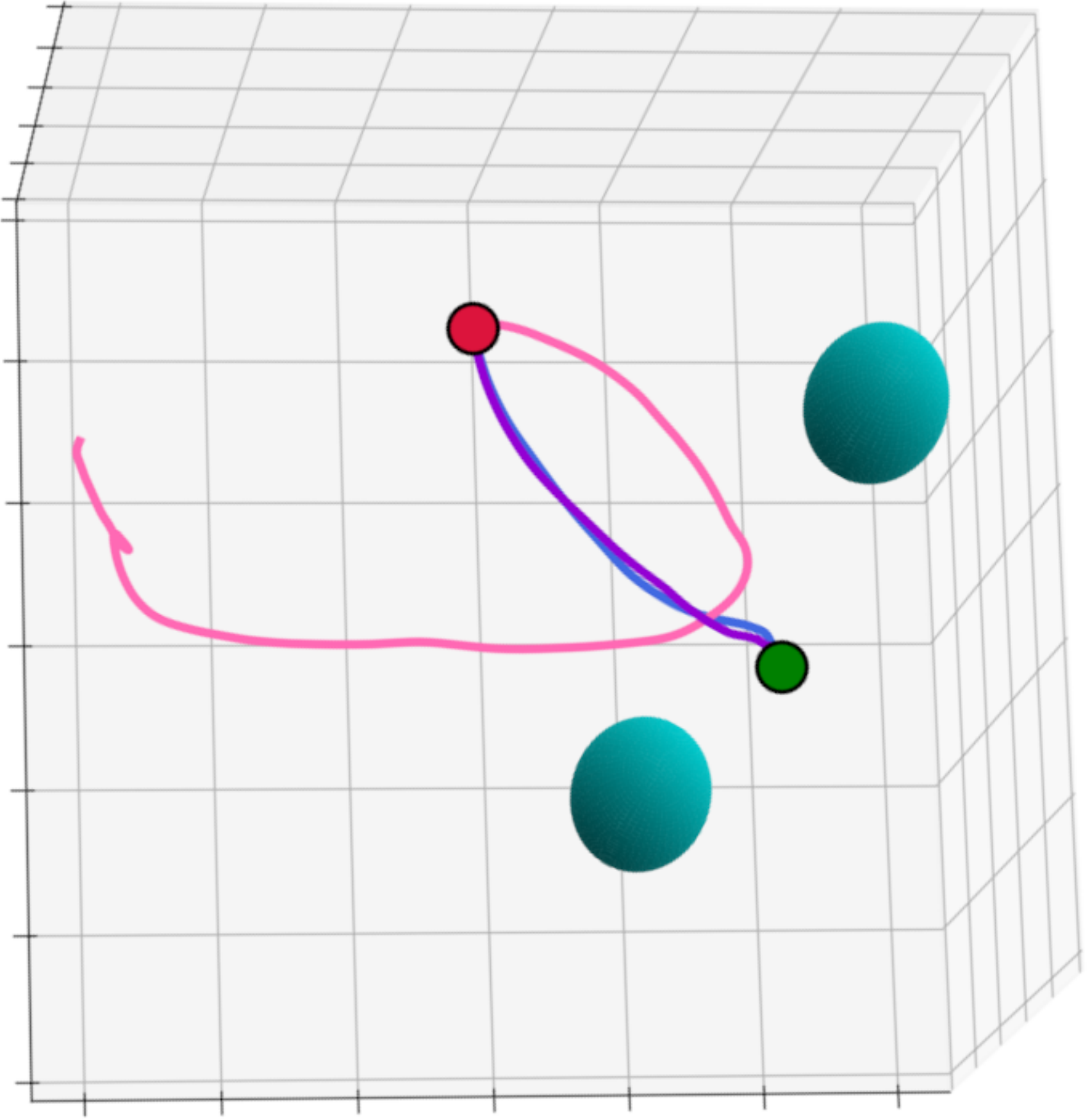}
\end{subfigure}
\hspace{1ex}
\begin{subfigure}[b]{0.25\textwidth}
    \centering
    \includegraphics[width=\textwidth]{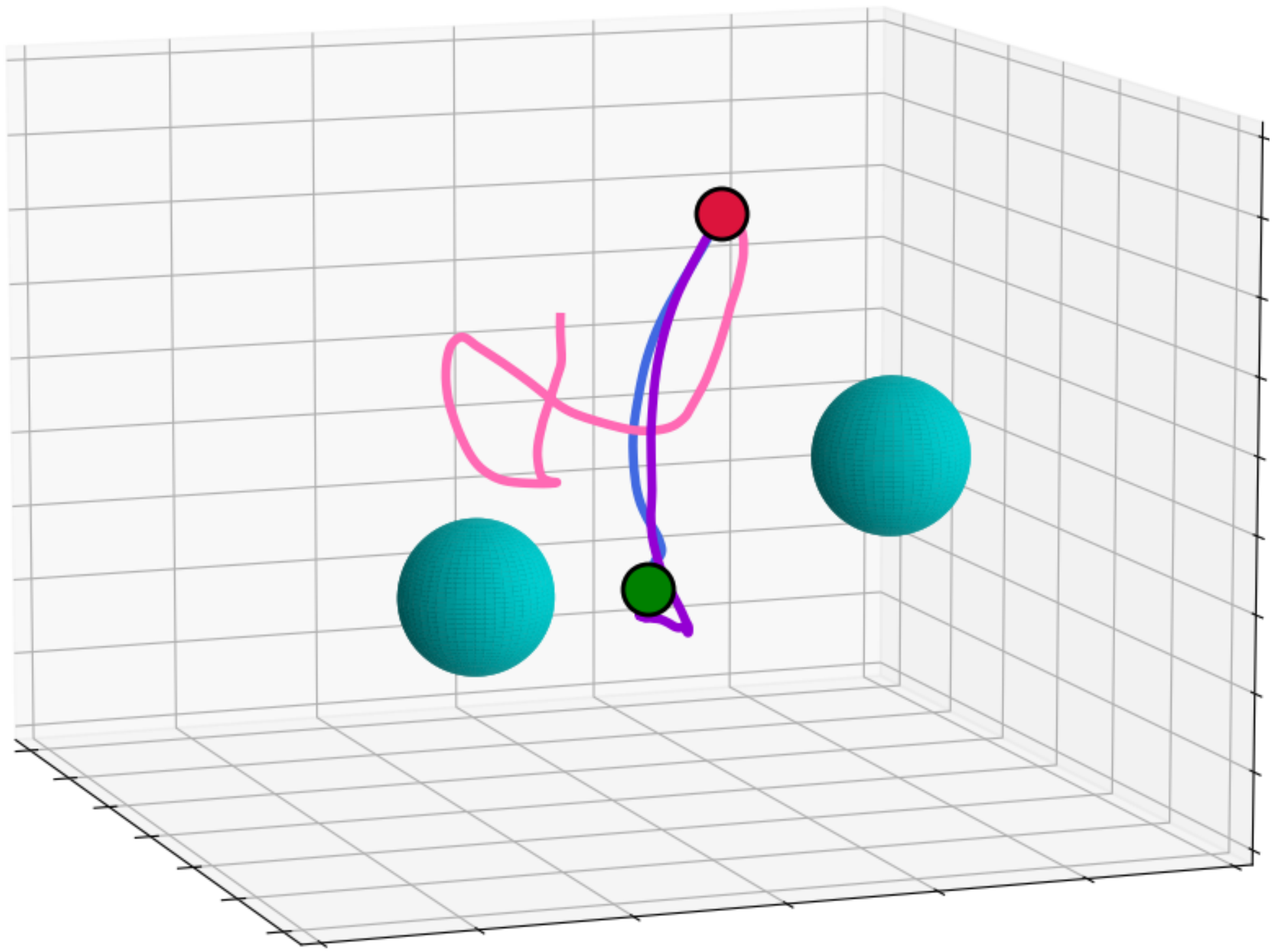}
\end{subfigure}
}
\caption{Trajectories of the Franka arm end effector when controlled by \textcolor{RoyalBlue}{\textbf{MPPI} with 512 samples (blue)}, \textcolor{CarnationPink}{\textbf{MPPI} with 16 samples (pink)}, and \textcolor{Fuchsia}{\textbf{L2O-MPC} with 16 samples (purple)} to move from the starting position (red) to the goal (green) while avoiding obstacles (cyan). Plots in the same row are from the same environment but viewed from differing perspectives.}
\label{fig:3d_plots}
\vspace{-3ex}
\end{figure}

\begin{table}[t]
\centering
\caption{Trajectory Statistics for \textsc{Franka Obstacles}.}
\begin{tabular}{c|cc}
\hrulethick
\textbf{\# Samples} & \textbf{Length} & \textbf{Avg. Jerk} \\
\hline
512 & 1.091 & 1.008\\
256 & 1.030 & 1.023\\
128 & 0.975 & 1.049\\
64 & 0.916 & 1.077\\
32 & 0.858 & 1.104\\
16 & 0.854 & 1.527\\
\hrulethick
\end{tabular}
\label{tab:stats}
\vspace{-2ex}
\end{table}

\Cref{fig:images} illustrates two different example trajectories of the Franka arm avoiding different amounts of obstacles.
These same environments are depicted in \Cref{fig:3d_plots}, where we show end effector trajectories under \textcolor{RoyalBlue}{\textbf{MPPI} with 512 samples}, \textcolor{CarnationPink}{\textbf{MPPI} with 16 samples}, and \textcolor{Fuchsia}{\textbf{L2O-MPC} with 16 samples}.
We can see that \textcolor{CarnationPink}{\textbf{MPPI} with 16 samples} quickly diverges and is unable to reach the goal.
On the other hand, \textcolor{Fuchsia}{\textbf{L2O-MPC} with 16 samples} is able to better make use of the samples and still reach the goal while avoiding all obstacles.
Moreover, the \textcolor{Fuchsia}{\textbf{L2O-MPC}} controller was trained only on environments that contain two obstacles.
Therefore, these results also indicate that the learned controller may generalize to novel environments on which it was not trained.

Qualitatively, the \textbf{L2O-MPC} trajectories appear to be slightly more jittery than the \textbf{MPPI} expert.
In \Cref{tab:stats}, we provide the average relative jerk between \textbf{L2O-MPC} and \textbf{MPPI} for successful test runs at different sample counts.
Indeed, we see that the \textbf{L2O-MPC} trajectories are less smooth than those of \textbf{MPPI}, and this effect is exacerbated at lower sample counts. 
Additionally, we provide the average relative trajectory length across all test environments.
With more samples, \textbf{L2O-MPC} has slightly longer trajectories than \textbf{MPPI}, indicating that it is slower at reaching the goal.
However, when given access to fewer samples, \textbf{L2O-MPC} consistently has shorter trajectories, as it more often reaches the goal.
Therefore, while \textbf{L2O-MPC} is often jerkier and sometimes slower than the expert with full samples, it succeeds more often in achieving the desired objective in a timely fashion than \textbf{MPPI} given the same number of samples.

\section{Conclusion}
We presented a method for improving upon standard sampling-based MPC algorithms by learning a better update rule.
This provides a novel way to incorporate learning into model-based control algorithms, which is orthogonal to the standard approaches of learning or fine-tuning the dynamics model and/or cost function.
We contend with noisy gradients by learning how to more effectively update the control distribution.
By using structured sampling strategies, we are able to provide more information to the learned update and better utilize fewer samples.
We show through empirical evaluations that our learned controllers remain competitive or outperform a baseline MPPI controller with access to the same number of samples.
%This represents a promising avenue for utilizing powerful control algorithms on resource-constrained systems.
This demonstrates the viability of the learning-to-optimize framework in the context of control, opening the door for a variety of techniques to be applied to improving the performance of optimization-based controllers and planners.
While we leveraged imitation learning to train the optimizers, this is just one possible option and an interesting direction for future work is to explore using reinforcement learning to see if it can outperform the expert and model-free methods.
Since performance of sampling-based methods relies so heavily on thorough exploration of the sample space, another possible avenue is to learn how to generate better samples in addition to better updates.

\bibliography{references}
\bibliographystyle{IEEEtran}

\end{document}